\documentclass[letterpaper]{IEEEtran}

\usepackage[utf8]{inputenc} 
\usepackage[T1]{fontenc}    
\usepackage{hyperref}       
\usepackage{url}            
\usepackage{booktabs}       
\usepackage{amsfonts}       
\usepackage{nicefrac}       
\usepackage{microtype}      
\usepackage{cite}

\usepackage{balance} 
\usepackage{multirow}
\usepackage{color,soul}	
\usepackage[font=footnotesize,labelfont=bf,skip=1pt]{caption}
\usepackage{subcaption}
\usepackage{bbm}
\usepackage{bm}
\usepackage{amsmath,amssymb}
\usepackage{amsmath}
\usepackage{cleveref}
\usepackage{wrapfig}
\usepackage{float}

\usepackage{tikz}
\usepackage{amsmath,amsthm,amssymb,amsfonts}
\usepackage{upgreek}
\usepackage{textgreek}
\usepackage{mathtools,mathdots}
\usepackage{xparse}
\usepackage{nameref}
\usepackage{graphicx}
\usepackage{subcaption}
\usepackage{float}
\usepackage{nicefrac}
\usepackage{multirow}
\usepackage{multicol}
\usepackage{hyperref}
\usepackage{multimedia}
\usepackage{arydshln}
\usepackage{substr}
\usepackage{mathrsfs}
\let\oldmathcal=\mathcal
\renewcommand{\mathcal}[1]{
    \IfSubStringInString{#1}{ABCDEFGHIJKLMNOPQRSTUVWXYZ}{\oldmathcal{#1}}{
    \IfSubStringInString{#1}{abcdefghijklmnopqrstuvwxyz}{\mathsf{#1}}{
    \ifthenelse{\equal{#1}{\epsilon}}{\textnormal{\straightepsilon}}{
    #1
    }}}
}

\usepackage{algorithm}
\usepackage{algpseudocode}

\algrenewcommand\alglinenumber[1]{#1:}

\usepackage{pbox}

\newcommand{\defined}{\vcentcolon=}

\newcommand{\reals}{\mathbb{R}}

\newcommand{\Mat}[1][]{\ifthenelse{\equal{#1}{}}{\text{Mat}}{\text{Mat}(#1)}}

\newcommand{\SE}[1]{\text{SE}(#1)}
\newcommand{\se}[1]{\mathfrak{se}(#1)}

\newcommand{\TSE}[2][]{
	\ifthenelse{\equal{#1}{}}
	{{T\SE{#2}}}
	{{T_{#1}\SE{#2}}}
}
\newcommand{\dualTSE}[2][]{
	\ifthenelse{\equal{#1}{}}
	{{T^*\SE{#2}}}
	{{T^*_{#1}\SE{#2}}}
}

\newcommand{\dif}{\mathrm{d}}

\newcommand{\set}[1]{\left\{#1\right\}}

\newcommand{\abs}[1]{\left|#1\right|}
\newcommand{\norm}[1]{\left\lVert#1\right\rVert}

\newcommand{\transpose}{\intercal}

\newcommand{\Trace}[1]{\text{Tr}\left(#1\right)}

\newcommand{\Frobenius}{\text{F}}
			
\newcommand{\Vector}[1]{#1}
\newcommand{\Matrix}[1]{\mathrm{#1}}

\newtheorem{proposition}{Proposition}
\newtheorem{remark}{Remark}

\usepackage{algorithm}
\usepackage{algpseudocode}

\algrenewcommand\alglinenumber[1]{#1:}

\newcommand{\position}{r}
\newcommand{\positions}{\Vector{r}}
\newcommand{\director}{\Vector{d}}
\newcommand{\orientation}{\Matrix{Q}}
\newcommand{\strain}{\epsilon}
\newcommand{\strains}{{\Vector{\strain}}}
\newcommand{\straintrajectory}{\mathcal{\strain}}
\newcommand{\curvatures}{\Vector{\kappa}}
\newcommand{\curvature}{\kappa}
\newcommand{\shears}{\Vector{\nu}}
\newcommand{\shear}{\nu}
\newcommand{\internalforce}{n}
\newcommand{\internalforces}{\Vector{n}}

\newcommand{\internalcouples}{\Vector{m}}

\newcommand{\posture}{\mathcal{q}}
\newcommand{\pose}{\Matrix{q}}
\newcommand{\twists}{\Matrix{\xi}}

\newcommand{\costatepose}{\Matrix{\lambda}}

\newcommand{\rigidity}{\Matrix{R}}

\newcommand{\symmetricpart}{\Matrix{M}}

\newcommand{\muscle}{\text{m}}

\newcommand{\data}{\text{d}}

\newcommand{\dataset}{\mathcal{D}}

\newcommand{\storedenergy}{W}
\newcommand{\potential}{\mathcal{V}}

\newcommand{\staticHamiltonian}{H}

\newcommand{\intrinsic}{\circ}

\newcommand{\decisionvariable}{u}
\newcommand{\decisionvariables}{\Vector{\decisionvariable}}
\newcommand{\decisionvariabletrajectory}{\mathcal{\decisionvariable}}

\newcommand{\musclepositions}[1][]{\positions^{\ifthenelse{\equal{#1}{}}{\muscle}{#1}}}
\newcommand{\musclerelativepositions}[1][]{\Vector{\gamma}^{\ifthenelse{\equal{#1}{}}{\muscle}{#1}}}

\newcommand{\cost}{\mathsf{J}}

\newcommand{\costatestrain}{\eta}
\newcommand{\costatestrains}{\Vector{\costatestrain}}

\newcommand{\musclelength}[1][]{\ell^{\ifthenelse{\equal{#1}{}}{\muscle}{#1}}}
\newcommand{\musclestrain}[1][]{\strain^{\ifthenelse{\equal{#1}{}}{\muscle}{#1}}}
\newcommand{\muscleshears}[1][]{\shears^{\ifthenelse{\equal{#1}{}}{\muscle}{#1}}}
\newcommand{\muscletangent}[1][]{\Vector{t}^{\ifthenelse{\equal{#1}{}}{\muscle}{#1}}}
\newcommand{\muscleforce}[1][]{\internalforce^{\ifthenelse{\equal{#1}{}}{\muscle}{#1}}}
\newcommand{\maxmusclestress}[1][]{\sigma^{\ifthenelse{\equal{#1}{}}{\muscle}{#1}}}
\newcommand{\maxmuscleforce}[1][]{\internalforce^{\ifthenelse{\equal{#1}{}}{\muscle}{#1}}_\text{max}}
\newcommand{\muscleforces}[1][]{\internalforces^{\ifthenelse{\equal{#1}{}}{\muscle}{#1}}}
\newcommand{\musclecouples}[1][]{\internalcouples^{\ifthenelse{\equal{#1}{}}{\muscle}{#1}}}
\newcommand{\muscleactivation}[1][]{a^{\ifthenelse{\equal{#1}{}}{\muscle}{#1}}}
\newcommand{\staticmuscleactivation}[1][]{\alpha^{\ifthenelse{\equal{#1}{}}{\muscle}{#1}}}

\newcommand{\musclestoredenergy}[1][]{W^{\ifthenelse{\equal{#1}{}}{\muscle}{#1}}}

\newcommand{\LM}[1][]{\text{LM}{\ifthenelse{\equal{#1}{}}{}{_{#1}}}}
\newcommand{\OM}[1][]{\text{OM}{\ifthenelse{\equal{#1}{}}{}{_{#1}}}}

\newcommand{\CameraDLTParam}{P}
\newcommand{\CameraIntensity}{I}
\newcommand{\ImageVelocity}{\Delta v}
\newcommand{\ImagePosition}{v}
\newcommand{\CameraCardinality}{i}

\usepackage{tikz}

\newcommand{\BibTeX}{\rm B\kern-.05em{\sc i\kern-.025em b}\kern-.08em\TeX}

\usepackage[normalem]{ulem}

\title{A physics-informed, vision-based method to reconstruct all deformation modes in slender bodies\vspace{-5pt}}

\author{
    \IEEEauthorblockN{Seung Hyun Kim\IEEEauthorrefmark{1}, Heng-Sheng Chang\IEEEauthorrefmark{1}, Chia-Hsien Shih\IEEEauthorrefmark{1}, Naveen Kumar Uppalapati\IEEEauthorrefmark{2}, Udit Halder\IEEEauthorrefmark{2},\\}
    \IEEEauthorblockN{Girish Krishnan\IEEEauthorrefmark{1}, Prashant G. Mehta\IEEEauthorrefmark{1}, Mattia Gazzola\IEEEauthorrefmark{1}\IEEEauthorrefmark{3}\vspace{-10pt}}
}

\begin{document}
\maketitle

\let\thefootnote\relax\footnote{\IEEEauthorrefmark{1}Mechanical Science and Engineering, University of Illinois at Urbana-Champaign}
\footnotetext[0]{\IEEEauthorrefmark{2}Coordinated Science Laboratory, University of Illinois at Urbana-Champaign}
\footnotetext[0]{\IEEEauthorrefmark{3}Corresponding Author: Mattia Gazzola, mgazzola@illinois.edu}
\footnotetext[0]{(https://github.com/GazzolaLab/BR2-vision-based-smoothing)}


\begin{abstract}
This paper is concerned with the problem of estimating (interpolating and smoothing) the shape (pose and the six modes of deformation) of a slender flexible body from multiple camera measurements. This problem is important in both biology, where slender, soft, and elastic structures are ubiquitously encountered across species, and in engineering, particularly in the area of soft robotics. The proposed mathematical formulation for shape estimation is physics-informed, based on the use of the special Cosserat rod theory whose equations encode slender body mechanics in the presence of bending, shearing, twisting and stretching. The approach is used to derive numerical algorithms which are experimentally demonstrated for fiber reinforced and cable-driven soft robot arms. These experimental demonstrations show that the methodology is accurate (<5 mm error, three times less than the arm diameter) and robust to noise and uncertainties.
\end{abstract}


\IEEEpeerreviewmaketitle
\section{Introduction}
Biological creatures that are slender or possess slender appendages, exploit the elasticity and compliance afforded by their bodies to perform and simplify a variety of tasks, from locomotion (snakes, eels, fishes \cite{Gray:1946,Domenici:1997,Goldman:2010,Gazzola:2015,Schiebel:2019,Thandiackal:2021}) to manipulation (octopuses, elephants, plants \cite{Kier:2007,Laschi:2012,Must2019}), and more generally to conform, adapt and respond to environmental interference. Bio-inspired slender structures are also being increasingly incorporated in engineering, and particularly in robotics, to enhance safety, dexterity and adaptivity \cite{Trivedi2008, Kim2013, Rus:2015}. Nonetheless, despite over a decade of soft robotic research, we have only begun to appreciate the inextricable nexus that exist between elasticity, control and environmental context. Thus, to support and advance biological discovery and engineering applications, as well as aid modeling and simulation efforts, there is a growing interest in accurate, robust, cost effective, and non-invasive technologies for shape and strains estimation in slender flexible bodies.


The problem of shape/strains estimation is complicated because elastic elements, whether biological or artificial, are subject to long-range stress propagation effects where all six modes of deformations (normal/binormal bending and shear, twist and stretch) can be simultaneously engaged. As a consequence, localized loads are communicated along the entire structure in a nonlinear fashion, leading to complex dynamics and global morphological reconfigurations \cite{Gazzola:2018,Charles:2019}. These are among the mechanisms credited to critically contribute to animals' superior agility, dexterity, and ability to cope with external factors, safely and robustly.
Consequently, they have important implications \cite{Pfeifer:2007} in terms of body architectural organization, actuation and control \cite{Kier:2007,Li:2012,Zhang:2019,chang2020energy,Naughton:2021,chang2021controlling}.
Hence the impending need, for correct mechanistic interpretation, of methods able to quantify from experiments \textit{all} continuum deformation, particularly in the presence of highly stretchable and shearable living or elastomeric materials \cite{Zhang:2019}. 

\begin{figure}[t]
\centering
\includegraphics[scale=0.34]{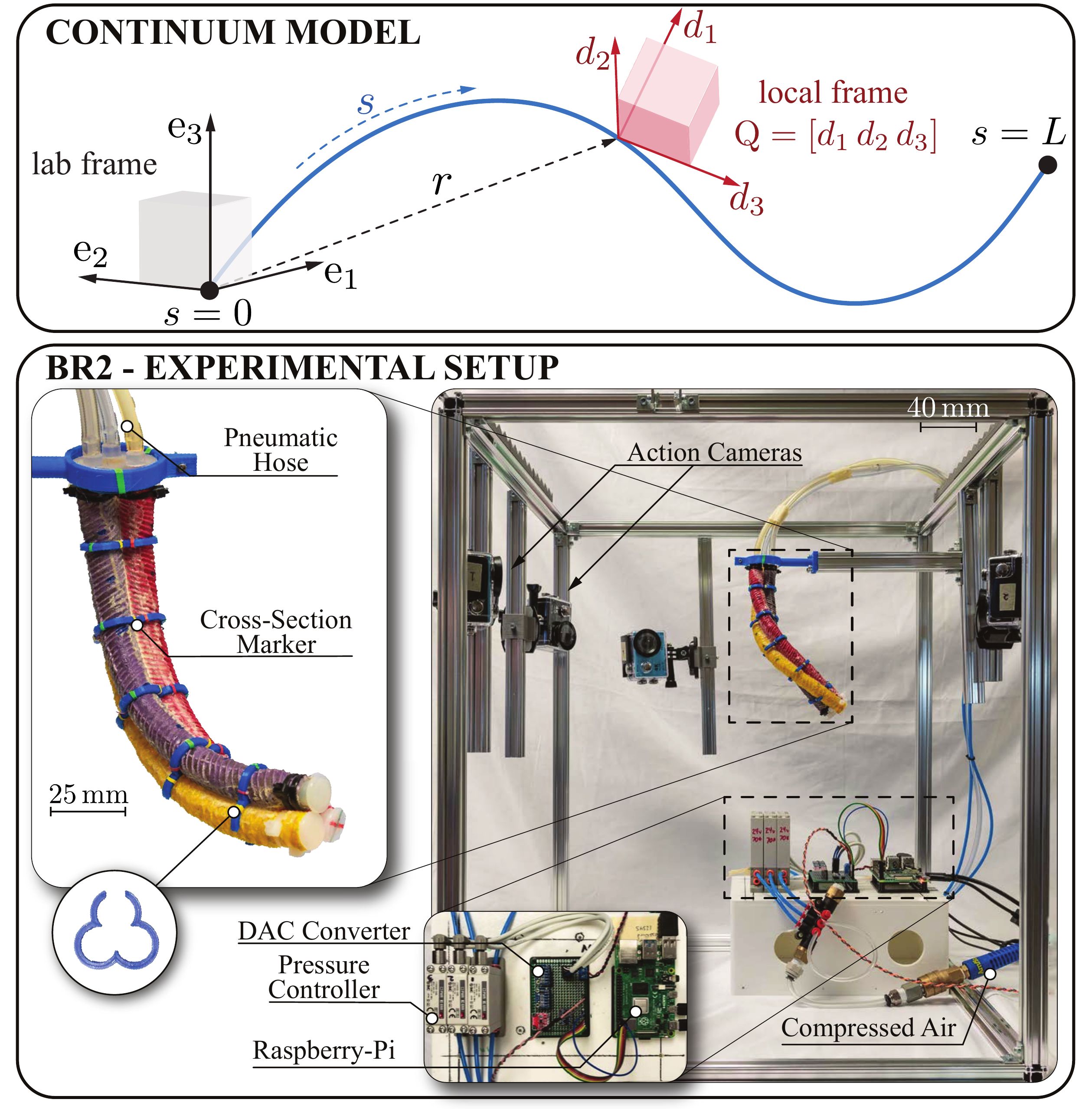}
\captionsetup{belowskip=-10pt}
\caption{Cosserat rod model $\&$ Experimental setup -- a soft BR$^2$ manipulator \cite{Uppalapati2021a} is integrated in our vision-tracking system. The BR$^2$ is constituted by three parallel elements, individually actuated by compressed air. Actuation signals are generated from Raspberry-Pi using ROS, and are relayed to SMC valves (ITV0031-2UBL) for actual pressure regulation.}
\vspace{-8pt}
\label{fig:1}
\end{figure}

The goal of this paper is to develop and demonstrate methods and algorithms to estimate, from multiple cameras images, the (continuum) bend, shear, twist and stretch strain functions along the longitudinal axis of a slender flexible body, in 3D space. The vision-based approach is selected here because it is versatile, relatively inexpensive, does not interfere with the system and can be deployed in a variety of conditions and environments.

The contributions of this paper are two-fold. The theoretical contribution is a novel optimization formulation to estimate {\em both} slender body's pose and its associated six strain functions from discrete measurements. The formulation is physics-informed, based on the use of Cosserat rod theory \cite{antman1995nonlinear} whose equations encode slender body mechanics in the presence of bending, shearing, twisting and stretching. A corresponding numerical solution algorithm is presented. The practical contribution lies in the experimental demonstration of the method for two soft robotic arms, one based on Fiber Reinforced Elastomeric Enclosures (FREE) \cite{Uppalapati2018,Singh2017,Bishop-Moser2013} and one cable-driven \cite{Laschi:2012,Wang2017,Bern2019}.
The methodology is shown to be scalable, accurate (<5 mm error, three times less than the arm diameter), and robust to noise and uncertainties in a variety of conditions, paving the way for novel analyses of soft systems, and improved control.

\medskip
\noindent \textbf{Related work}. Apart from the vision-based approach of this paper, there are a number of continuum sensing techniques, such as Fiber Bragg Grating \cite{Xu2016}, electromagnetic sensors \cite{FernandezGuzman2020, Bergamini2014} or liquid metal sensors \cite{Cho2021,Morrow2016,Park2013}, all of which are effective but may be expensive, fabrication intensive, poorly scalable, and may interfere with body dynamics.

In vision-based experimental setups, whether individual/multiple standard, wide-angle monocular, or depth cameras are used, the reconstruction of slender body centerlines, orientations and strains relies on image data extraction and interpolation.
A variety of methods exist for point extraction (often facilitated by body markers), from direct linear transformation \cite{AbdelAziz1971} and optical flow \cite{Fragkiadaki2013} to clustering \cite{Reiter2011}, skeletonization \cite{Fan2020,Inoue2021}, and deep learning \cite{Mathis2018}.
Discrete centerline points can then be interpolated into a continuous shape, from which local curvatures and elongations are directly estimated.
If no additional information is employed, only bending and stretch can be estimated.
Off-centerline discrete points can provide discrete orientations, which can be used to estimate all strain functions.
Nonetheless, the direct numerical evaluation of bending, twist, stretch and (particularly) shear is sensitive, and interpolation quality is key. While machine learning \cite{Scharff2021,Reiter2011}, gradient-based methods \cite{Manakov2021}, or clothoid functions \cite{Fan2020} have improved over unreliable, standard B-spline techniques, challenges remain.
As a consequence, these reconstruction approaches typically focus on a subset of deformations whereby (combinations of) twist, stretch as shear are routinely neglected.


In the context of our work, two papers are of particular significance. AlBeladi \textit{et al.}~\cite{AlBeladi2020} propose a least-squares type smoothing problem with a parametrized kinematic model of curvature. While the approach was successfully applied to a specific soft robot arm \cite{Uppalapati2021a}, shear and stretch were not considered, and only in-plane bending was actually experimentally demonstrated. Fu \textit{et al.} \cite{Fu2021} use instead a Cosserat rod model for piece-wise reconstruction of a snake's shape. Although both of these papers \cite{AlBeladi2020,Fu2021} are related to our work, important differences exist. We discussed them as part of Remarks 1 and 2 in the methodological section of this paper. 

\medskip
\noindent \textbf{Outline.} The remainder of this paper is structured as follows: The problem statement and its mathematical abstraction appear in Sec.~II. Its solution and associated numerical algorithm is described in Sec.~III. Experimental setup appears in Sec.~IV and results are presented in Sec.~V. Finally, Sec.~VI contains conclusions and future directions.

\section{Problem Formulation}
\noindent \textbf{Setup and data collection.} The setup is depicted in Fig.~\ref{fig:1}. The soft arm is fixed at the base and is of nominal length $L$. The arc-length parameter $s\in[0,L]$ is used to parameterize the centerline of the arm, with $s=0$ at the base and $s=L$ at the tip. For the purposes of mathematical modeling, an inertial reference frame $\set{\text{e}_1,\text{e}_2,\text{e}_3}$ is affixed at the base of the arm. With respect to this frame, the \emph{pose} of the arm at $s$ is denoted
\begin{equation}
    \pose(s):=\begin{bmatrix}\orientation(s)&\positions(s)\\0&1\end{bmatrix}\in\SE{3}
\end{equation}
where $\positions(s)\in\reals^3$ is the position vector, and $\orientation(s)=\begin{bmatrix}\director_1(s)&\director_2(s)&\director_3(s)\end{bmatrix}$ is the orientation matrix. The orthonormal vectors $\{\director_1(s),~\director_2(s),~\director_3(s)\}$ are referred to as directors. As depicted in Fig.~\ref{fig:1}, the normal director $\director_1(s)$ and the bi-normal director $\director_2(s)$ span the cross section at location $s$ and $\director_3(s)=\director_1(s)\times\director_2(s)$.
It is convenient to interpret $\pose(s)$ as an element of the special Euclidean group $\SE{3}$. The associated Lie algebra is denoted as $\se{3}$.

A number $N_\data$ of discrete colored marker are mounted at known fixed locations along the arm, with the $j^\text{th}$ one at $s=s_j$ and $0=s_0<\cdots<s_{N_\data}=L$. By observing these markers using multiple cameras (Sec. \ref{sec:Experimental_Setup}), pose measurement data is obtained as follows
\begin{equation}
    \dataset\defined\set{(s_j,\pose_j):j=1,2,\cdots,N_\data}
\end{equation}
where the measured pose $\pose_j\in\SE{3}$ represents a noisy measurement of the true pose $\pose(s_j)$. 

For a given data set $\dataset$, the \emph{smoothing problem} is to obtain a smooth posture $\posture=\set{\pose(s)\in\SE{3}:0\leq s\leq L}$ of the arm.

\medskip
\noindent \textbf{Cosserat rod model of a soft arm.}
\begin{subequations}\label{eq:kinematics}
    The posture $\posture$ is modeled as a solution of an ordinary differential equation (ODE)
    \begin{equation}\label{eq:pose_kin}
      \tfrac{\dif\pose}{\dif s}(s)=\pose(s)\twists(s),\quad\pose(0)=\pose_0
    \end{equation}
    where $\pose_0$, known and fixed, is the pose  at the base of the arm, and the matrix $\twists$ is parameterized by strains $\strains$ as follows
    \begin{equation}
        \twists(s)=
        \begin{bmatrix}
            [\curvatures]^\times&\shears\\
            0&0
        \end{bmatrix}\in\se{3},\quad
        \strains(s)=\begin{bmatrix}\curvatures(s)\\\shears(s)\end{bmatrix}
    \end{equation}
    with $\curvatures=[\curvature_1 \; \curvature_2 \; \curvature_3]$ being bending/twist strains, and 
    $\shears = [\shear_1 \; \shear_2 \; \shear_3]$ shear/stretch strains. The operator $[\cdot]^\times$ is defined as
    \begin{equation*}
        \begin{bmatrix}\curvature_1\\\curvature_2\\\curvature_3\end{bmatrix}^\times\defined
        \begin{bmatrix}
            0&-\curvature_3&~~\curvature_2\\
            ~~\curvature_3&0&-\curvature_1\\
            -\curvature_2&~~\curvature_1&0
        \end{bmatrix}
    \end{equation*}

\end{subequations}
\begin{subequations}
    The ODE~\eqref{eq:pose_kin} represents the kinematic constraint, so that 
    given the strain $\straintrajectory=\set{\epsilon(s):0\leq s\leq L}$, $\posture$ is obtained simply by integrating~\eqref{eq:pose_kin}. In the special theory of Cosserat rods, a model for strains is indirectly specified by introducing the potential energy
    \begin{equation}
        \potential=\int_0^L \storedenergy(s,\strains(s))~\dif s
    \end{equation}
    where $\storedenergy:[0,L]\times\reals^6\to\reals$ is referred to as the {\em stored energy function}. In this paper, we adopt a linear elasticity model specified by the quadratic choice for the stored energy function
    \begin{equation}
        \storedenergy(s,\strains)=\tfrac{1}{2}\abs{\strains-\strains^\intrinsic(s)}_\rigidity^2
    \end{equation}
    where $\rigidity=\rigidity(s)\succ 0$ is the rigidity matrix, $\strains^\intrinsic(s)$ is the intrinsic strain, and $\abs{x}_{\Matrix{R}}\defined\sqrt{\Vector{x}^\transpose\Matrix{R}\Vector{x}}$ is the weighted norm.
\end{subequations}

\begin{remark}
A static equilibrium of a Cosserat rod (with a free boundary condition at $s=L$) is any extremizing solution of the following optimization problem
\begin{subequations}\label{eq:static_optimization_problem}
    \begin{align}
        \min_{\straintrajectory}&\quad\potential\\
        \textnormal{subject to}&\quad \tfrac{\dif\pose}{\dif s}=\pose\twists,\quad\pose(0)=\pose_0
    \end{align}
\end{subequations}
This way, the strain $\straintrajectory$ may be regarded as a decision variable (control). The optimization viewpoint has several advantages, such as stability or uniqueness, as described in \cite{bretl2014quasi,till2017elastic}.

In the presence of measurements $\dataset$, this optimization framework is readily extended to obtain the piece-wise pose $\set{q(s):s_{j-1}\leq s \leq s_j}$, by specifying the fixed-fixed boundary conditions $q(s_{j-1})=q_{j-1}$ and $q(s_{j})=q_j$. Such an approach is taken in~\cite{Fu2021}. 

In fact, the optimization formulation is not even necessary to obtain static equilibria.
For general types of storage functions and boundary conditions (including the fixed-fixed case), there are well-established, efficient numerical algorithms to obtain the static equilibria of the Cosserat rod~\cite{healey2005straightforward}.
\end{remark}

\medskip
There are several issues with the piece-wise approach: (i) the measurements are noisy so fixed-fixed boundary conditions may not be appropriate; (ii) because local changes in the potential energy affect the stresses in the entire rod, it may not be appropriate to assume that the piece-wise segments are independent; and (iii) the piece-wise approach will in general yield solutions where strains are discontinuous at the boundaries ($s=s_j$). Such solutions may not be physically realistic. 

Our objective in this paper is to modify the optimization problem \eqref{eq:static_optimization_problem} to assimilate the data $\dataset$ in a global fashion. This is done through a novel specification of the decision variables (control), objective function and constraints.

\medskip
\noindent \textbf{Optimization problem.}
In this paper, we propose 
$\decisionvariabletrajectory=\set{\decisionvariables(s)\in\reals^6:0\leq s\leq L}$ as a decision variable (control) of the following form
\begin{equation}
    \tfrac{\dif\strains}{\dif s}(s)=\decisionvariables(s),\quad \strains(0)=\strains_0   
\end{equation}
where the strain $\strains_0$ at the base is fixed and known. Such a choice has the advantage of yielding strains that are continuous along $s$. 

The objective function is defined as
\begin{equation}
    \cost(\decisionvariabletrajectory;\dataset):=\potential+\underbrace{\frac{\chi_\decisionvariables}{2}\int_0^L\abs{\decisionvariables(s)}^2~\dif s}_{\text{(regularization)}}+\underbrace{\sum_{j=1}^{N_\data}\Phi(\pose(s_j);\pose_j)}_{\text{(smoothing cost)}}
\end{equation}
where $\Phi(\pose(s_j);\pose_j)=\tfrac{\chi_\orientation}{2}\norm{\orientation(s_j)-\orientation_j}_\Frobenius^2+\tfrac{\chi_\positions}{2}\abs{\positions(s_j)-\positions_j}^2$, $\norm{\cdot}_\Frobenius$ is the Frobenius norm, and $\chi_\decisionvariables,\chi_\orientation,\chi_\position>0$. Each of these costs is self-explanatory: (i) $\potential$ is the potential energy, modeling intrinsic elasticity, from the special Cosserat rod theory; (ii) the smoothing cost penalizes the prediction error, i.e. the deviation of the estimated pose $\pose(s_j)$ from the measured pose $\pose_j$ (such a choice is well-established in the least square theory for smoothing problems with noisy measurements); and (iii) the integral control cost is a regularization term to obtain a unique solution.

\begin{remark}
With a vision-based system, a perhaps more natural definition of the prediction error is the difference between the measurements in camera images, and the projection of the estimate to camera space. This can then be used to define the least squares smoothing cost. Such an approach is in fact taken in~\cite{AlBeladi2020} using a parameterized curvature model of the soft arm.
Extension of the proposed framework to handle this and more general forms of smoothing cost is possible and a subject of future work.
\end{remark}

In summary, the optimization problem reads as follows
\begin{subequations}\label{eq:smoothing_problem}
    \begin{align}
        \min_{\decisionvariabletrajectory}&\quad\cost(\decisionvariabletrajectory;\dataset)\\
        \text{subject to}&\quad \tfrac{\dif\pose}{\dif s}=\pose\twists,\quad\pose(0)=\pose_0\\
        &\quad \tfrac{\dif\strains}{\dif s}=\decisionvariables,\quad\strains(0)=\strains_0
    \end{align}
\end{subequations}

\section{Solution: Arm Reconstruction} 
\noindent \textbf{Necessary conditions.} The optimization problem \eqref{eq:smoothing_problem} is solved by using optimal control theory. 
For this purpose, the control Hamiltonian is defined as
\begin{equation}
    \begin{aligned}
        \staticHamiltonian(\pose,\strains,\costatepose,\costatestrains,&\decisionvariables;s)\\
        &:=\Trace{\costatepose^\transpose\pose\twists}+\costatestrains^\transpose\decisionvariables-\storedenergy(s,\strains)-\tfrac{\chi_\decisionvariables}{2}\abs{\decisionvariables}^2
    \end{aligned}
\end{equation}
where $(\pose,\strains)\in\SE{3}\times\reals^6$ is the state, $(\costatepose,\costatestrains)\in\dualTSE[\pose]{3}\times\reals^6$ is the co-state (the Lagrange multipliers associated with the constraints), $\decisionvariables\in\reals^6$ is the decision variable (control), and the arc-length parameter $s\in[0,L]$ is the independent coordinate. In order to write the Hamilton's equations, it is convenient to first express the co-state $\costatepose$ in terms of $(\internalcouples, \internalforces)$-coordinates
\begin{equation}\label{eq:costate_form}
    \costatepose=\begin{bmatrix}
        \tfrac{1}{2}\orientation\left([\internalcouples]^\times+\symmetricpart\right)&\orientation\internalforces\\
        0&0
    \end{bmatrix}
\end{equation}
where $\symmetricpart:=-\orientation^\transpose\left[\left(\orientation\internalforces\right)\positions^\transpose+\positions\left(\orientation\internalforces\right)^\transpose\right]\orientation$. In elasticity theory, these coordinates are the internal stresses, couple $\internalcouples$ and force $\internalforces$, both represented in the local frame. With a slight abuse of notation, the coordinate $(\internalcouples,\internalforces)\in\reals^3\times\reals^3$ is referred to as the co-state of pose. The form of $\costatepose$, as shown in \eqref{eq:costate_form}, is obtained through a direct computation of the corresponding Hamilton's equations, the proof of which is omitted here for brevity. The Hamilton's equations in the reduced coordinates are described next.
\begin{proposition} \label{prop:smoothing_problem_solution}
    Consider the optimization problem \eqref{eq:smoothing_problem}. 
    Suppose $\decisionvariabletrajectory$ is the minimizer and $(\posture, \straintrajectory)$ is the corresponding state trajectory.
    Then there exists a co-state trajectory $(\internalcouples,\internalforces):\cup_{j=1}^{N_d} (s_{j-1},s_j)\to\reals^3\times\reals^3$ and $\costatestrains:[0,L]\to\reals^6$, all not equivalently zero, such that
    \begin{align}
        \tfrac{\dif\pose}{\dif s}&=\pose\twists,\quad\pose(0)=\pose_0\label{eq:pose_evolution}\\
        \tfrac{\dif\strains}{\dif s}&=\decisionvariables,\quad\strains(0)=\strains_0\label{eq:strains_evolution}
    \end{align}
    and
    \begin{align}
        \frac{\dif}{\dif s}\begin{bmatrix}\internalcouples\\\internalforces\end{bmatrix}&=-\begin{bmatrix}\curvatures\times\internalcouples+\shears\times\internalforces\\
            \curvatures\times\internalforces
        \end{bmatrix}\label{eq:smoothing_problem_reduced_costatepose_evolution}\\
        \frac{\dif\costatestrains}{\dif s}&=-\begin{bmatrix}\internalcouples\\\internalforces\end{bmatrix}+\rigidity\left(\strains-\strains^\intrinsic\right) \label{eq:smoothing_problem_reduced_costatestrains_evolution}
    \end{align}
    with jump conditions at the boundaries 
    \begin{subequations}\label{eq:smoothing_problem_reduced_costate_jump_condition}
        \begin{align}
            \internalcouples(s_j^-)&=\internalcouples(s_j^+)-\chi_\orientation{\normalfont\text{vec}}\left[\orientation^\transpose\orientation_j-\orientation_j^\transpose\orientation\right]_{s=s_j}\\
            \internalforces(s_j^-)&=\internalforces(s_j^+)-\chi_\positions\left.\orientation^\transpose\left(\positions-\positions_j\right)\right|_{s=s_j}\\
            \internalcouples(s_{N_\data}^+)&=\internalforces(s_{N_\data}^+)=0,~ \eta(s_{N_\data})=0
        \end{align}
    \end{subequations}
    where $(\internalcouples(s_j^+),\internalforces(s_j^+))$ for $j< N_\data$ are obtained by integrating the co-state evolution \eqref{eq:smoothing_problem_reduced_costatepose_evolution} for $(s_j,s_{j+1})$,
    and $\text{vec}[\cdot]$ is the inverse operator of $[\cdot]^\times$. The optimal control is then obtained through the point-wise maximization of control Hamiltonian
    \begin{align}
        \decisionvariables(s)=\arg\max_{a}\staticHamiltonian(\pose(s),\strains(s),\costatepose(s),\costatestrains(s),a;s)=\tfrac{\costatestrains(s)}{\chi_\decisionvariables}
    \end{align}
\end{proposition}

The proof of the proposition is based on an application of the Pontryagin's maximum principle~\cite{dey2014control, Dey2015}. It is omitted here on account of space.

\begin{remark}
While the state equations \eqref{eq:pose_evolution}, \eqref{eq:strains_evolution} are the kinematics of the rod, the (reduced) co-state equations are interpreted in terms of the elasticity of the rod. Equations \eqref{eq:smoothing_problem_reduced_costatepose_evolution} are the standard equations of static equilibrium of a Cosserat rod \cite{antman1995nonlinear}, and equations \eqref{eq:smoothing_problem_reduced_costatestrains_evolution} can be interpreted as the modified \textit{constitutive laws} of the rod. The jump conditions \eqref{eq:smoothing_problem_reduced_costate_jump_condition} arise on account of the smoothing cost \cite{dey2014control}. 
\end{remark}

\begin{figure*}[th]
\centering
\includegraphics[width=\textwidth]{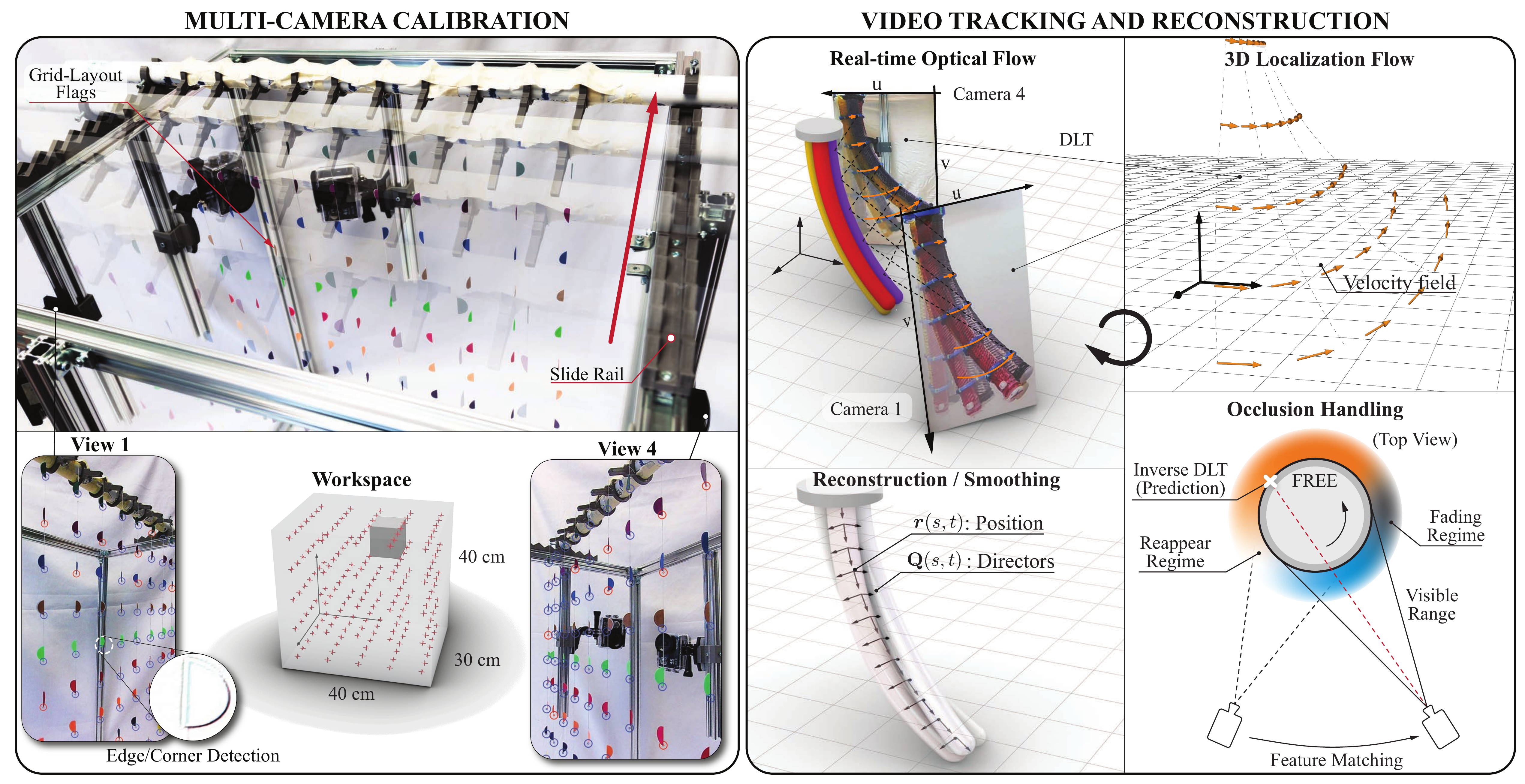}
\caption{
(Left) Camera calibration via multi-plane approach.
The tips of the flags, visible by all cameras, provide mapping between 3D-coordinates to 2D image spaces.
(Right)
Each camera tracks marker points on the soft arm using optical flow.
Advected marker locations are evaluated with DLT.
To handle occlusions, a feature matching technique is implemented to check the visibility of the markers by each camera.
Poses of tracking markers are then used in the optimization problem \eqref{eq:smoothing_problem} to reconstruct the soft arm’s posture.
}
\label{fig:2}
\vspace{-5pt}
\end{figure*}

\noindent \textbf{Algorithm.}
For a given data set $\dataset$, an iterative forward-backward algorithm (see also Sec. III-C in \cite{chang2020energy}) is used to numerically integrate the Hamilton's equations. During the $k^\text{th}$ iteration, the state trajectory $(\pose^{(k)}(s), \strains^{(k)}(s))$ is obtained by integrating \eqref{eq:pose_evolution}, \eqref{eq:strains_evolution} forward from the base to the tip; the co-state trajectory $(\internalcouples^{(k)}(s), \internalforces^{(k)}(s), \costatestrains^{(k)}(s))$ is obtained by integrating the co-state equations. Thus, for each segment $(s_{j-1},s_j)$, \eqref{eq:smoothing_problem_reduced_costatepose_evolution}, \eqref{eq:smoothing_problem_reduced_costatestrains_evolution} are integrated backward using boundary condition~\eqref{eq:smoothing_problem_reduced_costate_jump_condition} at $s=s_j$; finally, a gradient descent method is used to update the optimal decision variable
\begin{equation}\label{eq:update_law}
    \decisionvariables^{(k+1)}(s) = \decisionvariables^{(k)}(s) - \alpha\left(\decisionvariables^{(k)}(s)-\tfrac{\costatestrains^{(k)}(s)}{\chi_\decisionvariables}\right)
\end{equation}
where $\alpha>0$ is the update step size. Convergence results typically require sufficiently small values of $\alpha$. Pseudo code of the algorithm is shown in Algorithm \ref{alg:forward_backward}.

\begin{algorithm}[b]
    \caption{Solving the optimization problem \eqref{eq:smoothing_problem}}
    \label{alg:forward_backward}
    \begin{algorithmic}[1]
        \Require Data set $\dataset$ and state at base: $\pose(0)=\pose_0,~\strains(0)=\strains_0$
        \Ensure Optimal posture $\posture$ 
        \State Initialize: decision variable trajectory $\decisionvariables^{(0)}(s)$
        \For{$k=0$ to MaxIter}
            \State Update state $(\pose^{(k)}(s),~\strains^{(k)}(s))$
            \Statex \quad\quad by integrating \eqref{eq:pose_evolution}, \eqref{eq:strains_evolution} forward.
            \State Update co-state $(\internalcouples^{(k)}(s),~\internalforces^{(k)}(s),~\costatestrains^{(k)}(s))$ 
            \Statex \quad\quad by integrating \eqref{eq:smoothing_problem_reduced_costatepose_evolution}, \eqref{eq:smoothing_problem_reduced_costatestrains_evolution} backward
            \Statex \quad\quad with boundary condition \eqref{eq:smoothing_problem_reduced_costate_jump_condition}
            \State Update decision variables point-wise by using \eqref{eq:update_law}
        \EndFor
        \State Output the final posture $\posture$ by integrating \eqref{eq:pose_evolution}
        \Statex \quad with final strains $\strains^{(\text{MaxIter})}(s)$
    \end{algorithmic}
\end{algorithm}

\section{Experimental Setup}\label{sec:Experimental_Setup}
\begin{figure}[b!]
\centering
\includegraphics[scale=0.35]{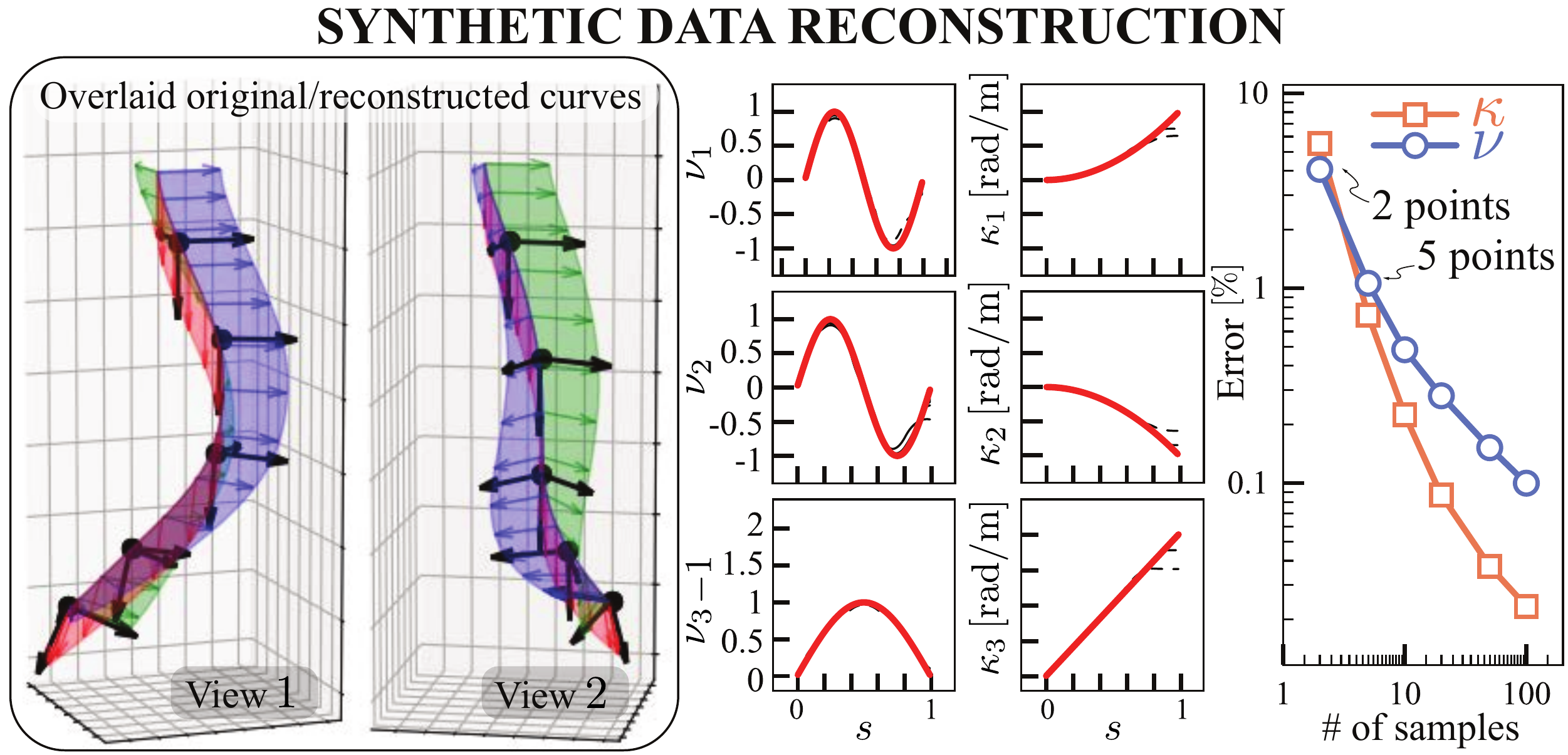}
\caption{
(Left) Two views show that the reconstructed pose ($\director_1$, green -- $\director_2$, blue -- $\director_3$, red) visually aligns with the 5 ground-truth samples (black).
(Middle)
The six strains include shear $\shear_1,\shear_2$, stretch $\shear_3-1$, bending $\curvature_1,\curvature_2$ and twisting $\curvature_3$.
Reconstruction of all strains (black) is plotted against ground-truth shear/stretch strains $\bar{\shears}$ and bending/twisting strains $\bar{\curvatures}$ (red)
(Right) Relative reconstruction error $e=\int_0^L \abs{\curvatures-\bar{\curvatures}}^2\dif s/(L\cdot\max_{s}\abs{\bar{\curvatures}})$ versus number of samples. Same error definition used for $\shear$.}
\label{fig:5}
\end{figure}

\noindent\textbf{The BR$^2$ soft continuum arm.} In order to demonstrate our physics-informed reconstruction method, we consider a slender and soft robotic arm integrated within a video-tracking environment.
We focus primarily on Fiber Reinforced Elastomeric Enclosures (FREEs \cite{Bishop-Moser2013}), due to their high degree of shape reconfigurability, versatility, cost effectiveness and overall promise for applications, from manipulation \cite{Uppalapati2018} to agriculture \cite{Uppalapati:2020}.
FREEs are pneumatic slender actuators embedding in their elastomeric shell inextensible fibers which, depending on the winding pattern and upon pressurization, lead to bending, twisting or elongation. We consider a BR$^2$\cite{Uppalapati2021a} architecture (Fig.~\ref{fig:1}) whereby three parallel FREEs, one bending and two twisting (clockwise and counterclockwise), are glued together.
By combining various primary deformation modes, the BR$^2$ attains complex morphologies and large workspaces.
It is thus an excellent candidate for testing the practical utility of our methods.
Further, the BR$^2$ can be extended by adding new sections in series (Fig.~\ref{fig:5}), increasing reconfigurability and workspace.
While our main interest is centered around the BR$^2$, we also consider a cable-driven soft-arm (Fig.~\ref{fig:5}), selected because of its propensity to shear (a mode we wish to capture).

\medskip
\noindent\textbf{Multi-camera environment.} Our video-tracking apparatus is constituted by an aluminum profile of size $40\times 30\times 40 \text{cm}$ that defines the data acquisition volume, and supports the arm as well as five AKASO EK7000 cameras (60fps, 1080p). The metal frame allows to strategically place the cameras, so that all marker points along the arm (Fig.~\ref{fig:2}) are visible by at least two cameras throughout any attainable motion.

Given this setup, marker positions in 3D space are obtained via standard Direct Linear Transformation (DLT \cite{AbdelAziz1971}), which processes 2D images from multiple cameras $\CameraCardinality=1,\dots,5$ to approximate 3D coordinates $(x,y,z)$
    \begin{align}
    \begin{split}
        \ImagePosition^i_1 &= \frac{\CameraDLTParam^i_1x+\CameraDLTParam^i_2y+\CameraDLTParam^i_3z+\CameraDLTParam^i_4}{\CameraDLTParam^i_9x+\CameraDLTParam^i_{10}y+\CameraDLTParam^i_{11}z+1} \\
        \ImagePosition^i_2 &= \frac{\CameraDLTParam^i_5x+\CameraDLTParam^i_6y+\CameraDLTParam^i_7z+\CameraDLTParam^i_8}{\CameraDLTParam^i_9x+\CameraDLTParam^i_{10}y+\CameraDLTParam^i_{11}z+1}
        \end{split}
    \end{align}
    \vspace{-8pt}
    \begin{align*}
        \begin{bmatrix}\ImagePosition^i_1\CameraDLTParam^i_9-\CameraDLTParam^i_1 & \ImagePosition^i_1\CameraDLTParam^i_{10}-\CameraDLTParam^i_2 & \ImagePosition^i_1\CameraDLTParam^i_{11}-\CameraDLTParam^i_3 \\ \ImagePosition^i_2\CameraDLTParam^i_9-\CameraDLTParam^i_5 & \ImagePosition^i_2\CameraDLTParam^i_{10}-\CameraDLTParam^i_6 & \ImagePosition^i_2\CameraDLTParam^i_{11}-\CameraDLTParam^i_7 \\ & \vdots & \end{bmatrix}
        \begin{bmatrix}x \\ y \\ z\end{bmatrix} = 
        \begin{bmatrix}\CameraDLTParam^i_4-\ImagePosition^i_1 \\ \CameraDLTParam^i_8-\ImagePosition^i_2 \\ \vdots \end{bmatrix}\label{eq:D\CameraDLTParamT}
    \end{align*}
where $\ImagePosition^i=(\ImagePosition_1^i,\ImagePosition_2^i)$ are the 2D coordinates in camera $i$ space, and $\CameraDLTParam^i=(\CameraDLTParam^i_1, \CameraDLTParam^i_2, \CameraDLTParam^i_3, ..., \CameraDLTParam^i_{11})$ are necessary calibration parameters. The above over-determined system is then solved via linear regression. We note that once $\CameraDLTParam^i$ is determined, DLT provides bidirectional mapping, which allows us to overlay reconstructed positions and orientations on the original image to visually assess the quality of the solution.

Camera calibration is critical and must be carried out thoroughly to achieve reliable accuracy.
While checkerboards in the lateral, vertical, and horizontal side planes are common calibration references, we noticed that interpolations in the enclosed volume result distorted.
To improve accuracy, we developed a scalable calibration method (Fig.~\ref{fig:2}) whereby a fine 3D reference grid is embedded throughout the workspace.
This is achieved via equispaced flags of alternating colors, hanging from a support rod along thin, transparent nylon wires, straightened by a weight at the bottom. The device provides a finely gridded yz-calibration plane, and a top slide rail allows to move the device to multiple x-positions, thus forming our volume calibration mesh. Typically, for calibration purposes, reference points of known $(x,y,z)$ coordinates are manually identified in 2D camera space to obtain accurate $(\ImagePosition_1^i,\ImagePosition_2^i)$-coordinates.
This is a time consuming and non-scalable procedure. To automate the process, after an initial manual identifications of a few (at least 12) points, we use the corresponding, minimally calibrated DLT to estimate each flag location in all 2D camera spaces, and then apply Harry's corner detection algorithm \cite{Harris1988} to precisely determine the $(\ImagePosition_1^i,\ImagePosition_2^i)$-coordinates of the flag's tips.
As a result, more than 1000 reference points, simultaneously visible by all cameras, could be collected, leading to a 2.6 mm maximum, and 1.4 mm average, calibration error throughout the entire workspace.

\medskip
\noindent\textbf{Video tracking.}
After calibration, for each camera recording, we manually select all reference marker points along the arm, and then trace them in time and space via the Lucas-Kanade's optical flow method \cite{Lucas1981}.
The algorithm assumes relatively constant object intensity $\CameraIntensity^i(\ImagePosition^i,t)=\CameraIntensity^i(\ImagePosition^i+\ImageVelocity^i,t+1)$ and considers the scalar advection $\nabla\CameraIntensity^i\cdot\ImageVelocity^i+\CameraIntensity^i_t=0$, where $\ImageVelocity^i=(\ImageVelocity^i_1,\ImageVelocity^i_2)$ is the optical flow. To recover $\ImageVelocity^i$, we employ a window of $19\times 19$ pixels (full marker size) across two consecutive frames, thus over-determining the advection system, which is solved via regression.


A challenge is posed by the frequent markers appearance and disappearance from camera view, due to arm occlusion.
Disappearance can be handled by monitoring the optical flow error which is related to the contrast gradient $\nabla\CameraIntensity$.
If the gradient magnitude in one direction is significantly larger than in any other direction, then the point is likely at the horizon, may be considered disappeared, and its tracking stops. Handling reappearances is more convoluted. While sophisticated pattern recognition techniques can be used, here we opt for a simpler inverse-DLT approach. If the 3D coordinate of a marker is known (because other cameras see it) then its position can be estimated in camera space, and colors and intensity spectrum of the marker (measured by cameras that see it) can be locally searched, to determine its reappearance.

\begin{figure*}[t!]
\centering
\includegraphics[width=\textwidth]{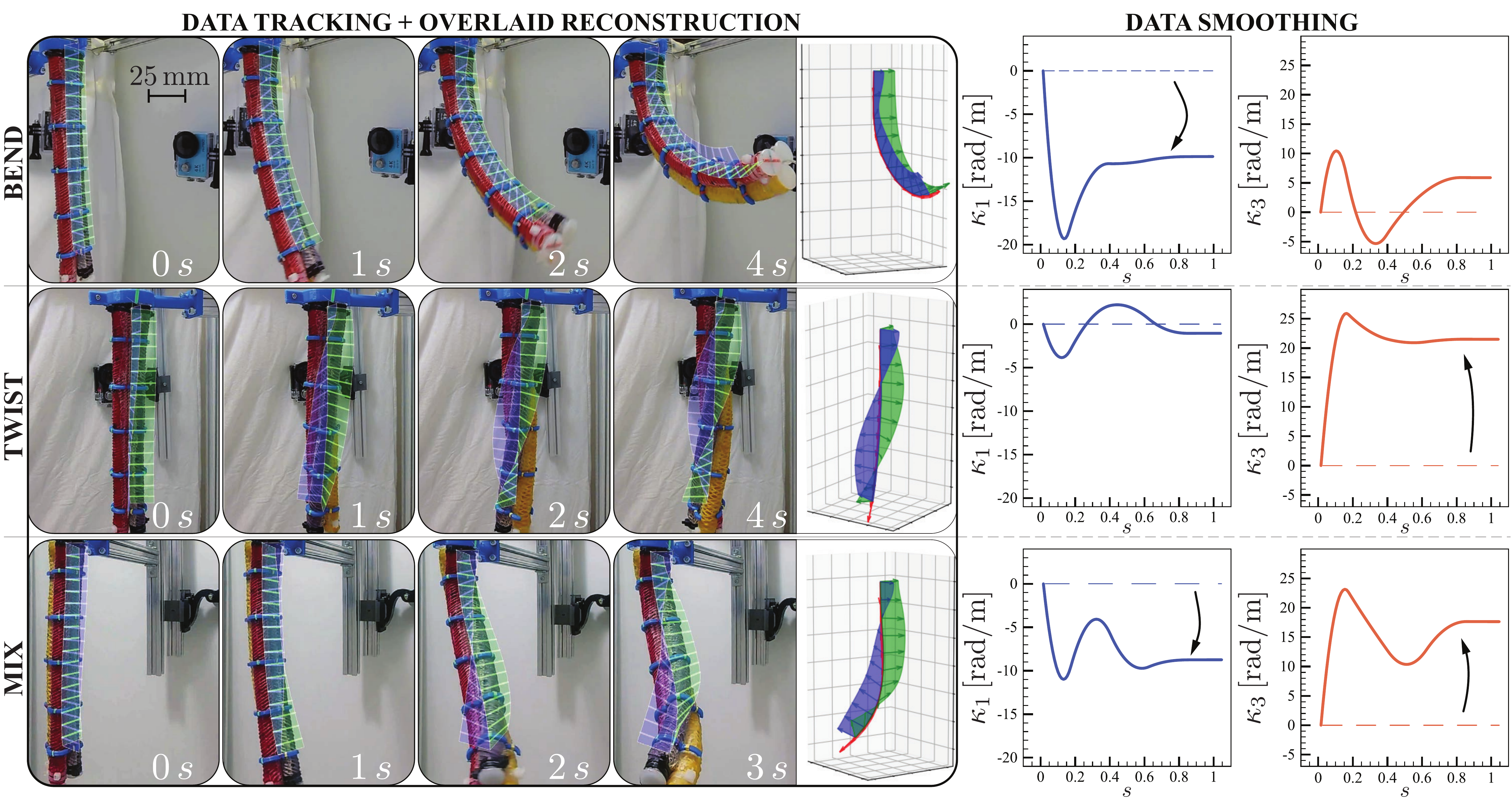}
\captionsetup{belowskip=-10pt}
\caption{
(Left) Reconstruction results are overlaid on top of BR$^2$ experimental recordings for bending ($35$ psi), twisting ($30$ psi), and mixed bending/twisting ($35$/$25$ psi) actuations.
(Right) Reconstructed bending $\curvature_1$ and twist $\curvature_3$ along the arm.
}
\label{fig:3}
\vspace{-5pt}
\end{figure*}

\section{Results and analysis}
A series of synthetic and robotic investigations is presented here to demonstrate the reconstruction abilities of our method, and its robustness to experimental disturbances.

\medskip
\noindent\textbf{Reconstruction from synthetic data.}
We start by generating a 3D curve based on six analytically known strain functions (Fig.~\ref{fig:5}).
These are arbitrarily determined, harmonic functions selected to produce strains of comparable magnitude.
The synthetic ground-truth curve is sampled along its arc-length at increasing resolution (from a minimum of 2 to a maximum of 100 points).
Corresponding position/orientation datasets are then used for reconstruction, to assess the impact of coarsening on the quality (relative $\mathcal{L}_2$-error norm) of the recovered strain functions versus the ground truth. As can be seen in Fig.~\ref{fig:5}, errors rapidly decrease reaching $\sim$1\% with 5 points. Even with just 2 points, reconstruction errors are within a reasonable $\sim$5\% level, demonstrating the applicability of this approach to sparse data.
This characterization was performed for a number of synthetic curves, consistently recovering similar trends.

\medskip
\noindent\textbf{Reconstruction of individual modes: BR$^2$ arm.}
Next, we move on to our first experimental demonstration (Fig. \ref{fig:3}).
Here, we employ a BR$^2$ arm and take advantage of its multi-modality to elicit individual or (simple) mixed deformation modes. This allows us to verify the consistency of reconstructed strain functions against the approximately known, intuitively expected BR$^2$ deformations.

The BR$^2$ is initially at rest in a straight, vertical configuration, with directors (i.e. marker orientations, five markers used) well aligned with camera views. Then, three experiments are performed in which only bending or twisting or both are activated. For all these cases, the arm motion is tracked and $\curvature_1$ (bending), $\curvature_3$ (twisting) curvature profiles are reconstructed.

As can be seen in the bending experiment of Fig.~\ref{fig:3}, $\curvature_1$ is most significant implying that the bending is approximately in plane, as expected. For twisting only, $\curvature_3$ is predominant and as a result the arm centerline remains straight, while individual FREEs helicoidally reshape around it. In both cases, the less significant modes are not exactly zero. This is because glueing, FREEs kinematic constraints and gravity give rise to residual or undesired stresses. A careful fabrication can mitigate such effects, although they cannot be entirely removed. Finally, both bending and twisting are simultaneously activated which, consistently, is reflected by the comparable magnitudes of $\curvature_1$ and $\curvature_3$. In all cases reconstructed centerlines and orientations are mapped back to camera space and overlaid on top of the arm, illustrating good visual agreement.

In these demonstrations we do not have known ground-truth curves to compare with. Then, we determine the reconstruction error relative to the arm tip position, which in turn is measured via an electromagnetic sensor (Patriot SEU, Polhemus, < 1mm uncertainty) glued to it.
We note that the error distance to the tip reflects the global error of our entire methodology, from calibration of reconstruction. Our system nonetheless was able to consistently recover the actual tip position within 5~mm error, which is less than one third of the BR$^2$ diameter.

\begin{figure*}[h!]
\centering
\includegraphics[width=\textwidth]{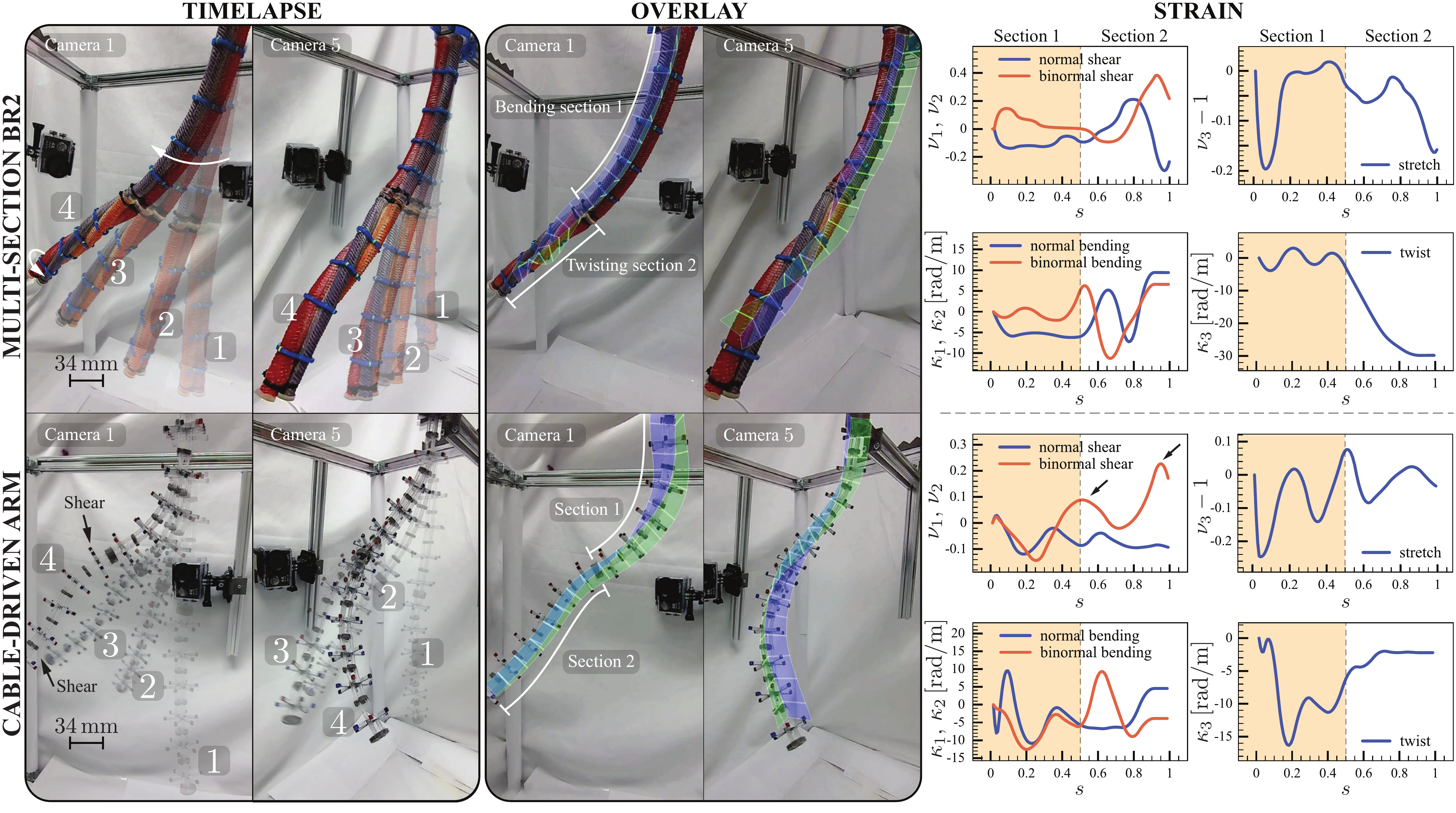}
\captionsetup{belowskip=-10pt}
\caption{
Study of multi-section BR$^2$ and a cable-driven arms.
(Left) Timelapse.
The multi-section BR$^2$ performs bending in the upper section and twisting in the lower section.
The cable-driven arm is bending both sections in two different directions and experiences twisting and shearing due to gravity.
(Middle) Reconstructed poses of final arms' configurations are overlaid.
(Right) 
Corresponding six continuous strain functions.
}
\label{fig:4}
\vspace{-5pt}
\end{figure*}

\medskip
\noindent\textbf{Simultaneous reconstruction of all modes: multi-section BR$^2$ and cable-driven arms.}
Here we consider a multi-section BR$^2$ arm, made of individual BR$^2$ serially stacked together, and a cable-driven arm (Fig.~\ref{fig:4}). The goal is to challenge our approach to simultaneously reconstruct all modes of deformation from technologically diverse and highly reconfigurable robots.

In the first experiment, the multi-section BR$^2$ is initially at rest in a straight configuration.
Upon actuation, section 1 is given a bending signal while section 2 receives a twisting one, realizing the reconfiguration sequence of Fig.~\ref{fig:4}.
By tracking the 11 markers along the arm, we reconstruct all strain functions.
As can be seen, in the first section $\curvature_1$ is predominant as expected, while in the second section $\curvature_3$ is most significant. All other modes are excited, albeit to a lesser extent, in a non-intuitive, non-linear fashion on account of the coupling between FREEs, and gravity.

In the second experiment, the cable-driven arm (made of two sections, 12 markers) is actuated so as to excite $\curvature_1$ in section 1 and $\curvature_2$ in section 2, as captured by the reconstruction. Further, due to its materials (softer than BR$^2$) and design (cables can render cross-sectional holders no longer perpendicular to the centerline), the arm is particularly susceptible to twisting and shearing effects, exacerbated by gravity. These expected additional deformations are apparent from camera images, and indeed are captured in our reconstruction (Fig.~\ref{fig:4}).

Finally, to further illustrate our method robustness to noise, we digitally perturbed the 12 markers positions along the cable-driven arm with uniform noise of up to 5 mm ($100\%$ of the overall system reconstruction error). Then, we recovered the new strains and compared them with the ones of Fig.~\ref{fig:4}, obtaining an average $\sim 10\%$ relative $\mathcal{L}_2$-error, confirming the ability of our approach to recover sensible physical solutions in the face of significant data disruption.

\section{Conclusion}
To fully understand how compliance mediates and assists control both in the biological and engineering domain, methods that are able to accurately quantify all modes of deformation in ubiquitous slender structures are necessary.
To respond to this need, we theoretically establish a physics-informed framework that reconstructs normal and binormal bending, shear, twist, and stretch strain functions along any generic slender structure.
Our approach relies on Cosserat rod theory, and seeks to obtain globally smooth strain functions so that the corresponding 3D shape minimizes both elastic energy and distance from discrete pose data.
The methodology is applied to reconstruct shapes and strains from multi-camera images of highly reconfigurable soft robotics arms (BR$^2$, multi-section BR$^2$, cable-driven).
Results demonstrate the generality, accuracy and robustness of our overall reconstruction setup, in the face of limited and noisy data.
Robotics demonstrations further underscore the practical utility of our integrated platform, for the future development of autonomous abilities in applications ranging from agroforestry \cite{Chowdhary:2019} to surgery and assistive care.


\bibliographystyle{IEEEtran}
\bibliography{Bib_Seung,Bib_HSChang,Bib_Cathy,Bib_Naveen, DaBib}



\end{document}